\let\oldvec\vec
\documentclass[runningheads]{article}
\let\vec\oldvec

\usepackage{arxiv}

\usepackage{nicefrac}       % compact symbols for 1/2, etc.
\usepackage{microtype}      % microtypography
\usepackage{lipsum}
\usepackage{xcolor}
\usepackage{cite}
\usepackage[utf8]{inputenc} % allow utf-8 input
\usepackage[T1]{fontenc}    % use 8-bit T1 fonts
\usepackage[bookmarks=false, unicode, colorlinks=true, urlcolor={blue!50!black}, linkcolor={red!50!black}, citecolor={green!50!black}]{hyperref}       % hyperlinks
\usepackage{url}            % simple URL typesetting
\usepackage{booktabs}       % professional-quality tables
\usepackage{nicefrac}       % compact symbols for 1/2, etc.
\usepackage{microtype}      % microtypography
\usepackage{lipsum}
\usepackage{graphicx}
\graphicspath{ {./ISVC19-RemoteSensing/images/} }
\usepackage{subfigure}
\usepackage{amsfonts, amsmath, amssymb} % For math fonts, symbols and environments
\usepackage{booktabs} % Top and bottom rules for table
\usepackage{wrapfig} % Allows in-line images
\usepackage{sidecap}
\sidecaptionvpos{figure}{t}
\usepackage{indentfirst}
\usepackage[vlined,ruled]{algorithm2e}

\usepackage{array}
\newcolumntype{L}[1]{>{\raggedright\let\newline\\\arraybackslash\hspace{0pt}}p{#1}}
\newcolumntype{C}[1]{>{\centering\let\newline\\\arraybackslash\hspace{0pt}}p{#1}}
\newcolumntype{R}[1]{>{\raggedleft\let\newline\\\arraybackslash\hspace{0pt}}p{#1}}

\usepackage{enumitem}
\usepackage{multirow}
\setlist[description]{leftmargin=8pt,labelindent=0pt,itemsep=0pt}
\setlist[itemize]{itemsep=0pt,parsep=0pt}
\setlist[enumerate]{itemsep=0pt,parsep=0pt, topsep=0pt}

\title{An Automatic Digital Terrain Generation Technique for Terrestrial Sensing and Virtual Reality Applications}
\author{
  Lee Easson\\
    Department of Computer Science and Engineering\\
    University of Nevada, Reno\\
    Reno NV 89557
   \And
 Alireza Tavakkoli \\
     Department of Computer Science and Engineering\\
    University of Nevada, Reno\\
    Reno NV 89557
 \And
 Jonathan Greenberg\\
    Department of Natural Resources and Environmental Science\\
    University of Nevada, Reno\\
    Reno NV 89557  
}

\begin{document}

\maketitle              % typeset the header of the contribution
\begin{abstract}
The identification and modeling of the terrain from point cloud data is an important component of Terrestrial Remote Sensing (TRS) applications. The main focus in terrain modeling is capturing details of complex geological features of landforms. Traditional terrain modeling approaches rely on the user to exert control over terrain features. However, relying on the user input to manually develop the digital terrain becomes intractable when considering the amount of data generated by new remote sensing systems capable of producing massive aerial and ground-based point clouds from scanned environments. This article provides a novel terrain modeling technique capable of automatically generating accurate and physically realistic Digital Terrain Models (DTM) from a variety of point cloud data. The proposed method runs efficiently on large-scale point cloud data with real-time performance over large segments of terrestrial landforms. Moreover, generated digital models are designed to effectively render within a Virtual Reality (VR) environment in real time.  The paper concludes with an in‐depth discussion of possible research directions and outstanding technical and scientific challenges to improve the proposed approach.

\keywords{Digital Terrain Model  \and Terrestrial Remote Sensing \and Geological Landmass Modeling.}
\end{abstract}

\section{Introduction}\label{sec:Intro}
Terrains are among the most fundamental features in any virtual application simulating a landmass, ranging from computer games to geological simulations.  For example, in an open-world massively multiplayer online role playing game, large‐scale natural environments maybe designed for players to explore, where a vast terrain is usually the first part of the authoring pipeline to be subsequently augmented with props that represent rocks, trees, plants and buildings. On the other hand, real‐world terrain are usually more complex and varied and may include plains, mountain ranges, and eroded valleys in a single environment. Terrain formation is a combination of long‐term and complex geological events with complicated physical and geological interactions amongst different components comprising the landmass. In addition, different geological features are dominant at difference range scales. These complexities contribute to many unsolved challenges in terrain modeling.

One definition of Digital Terrain Models (DTM), \cite{el2005digital}, relates to geometrical aspects of the 3D environment acquired from Laser scanning and is a continuous function mapping a 2D position $(x,y)$ to the terrain elevation $z=f(x,y)$. In this definition, the terrain is defined as the boundary between the ground and the air. Yet, there are certain geographical feature, such as overhangs \cite{pfeifer2005subdivision}, ground vegetation \cite{naesset2015vertical}, and large man-made structures, that may render the assumptions required for this definition inaccurate \cite{shan2018topographic}.

The aforementioned DTM will require utilizing a large amount of data collected by aerial or ground-based Laser scanning technology. This data is generally combined to produce a collection of points referred to as Point Clouds (PC). In essence, each point in a point cloud represented a location in the world from which the light emitted from the scanner is reflected back. The massive amount of data within even a small scanned region makes it necessary to represent the DTM using a more efficient structure. 

Several data structures are utilized in the literature that represent DTMs with varied levels of performance \cite{shan2018topographic}. These structures range from pixel-level representation of the elevation data by quantizing the 2D planimetric locations of the point clouds to a hybrid approach by interpolating points on the surface of a grid-mesh structure \cite{ackermann2004grid}. To improve the quality of the structure of the DTM, and with the popularity of triangulation techniques in computer graphics, several Triangulated Irregular Networks (TINs) are proposed with the goal of improving storage efficiency of the point cloud representation of the DTM \cite{axelsson2000generation}, with recent attempts to improve the performance of the triangulation approaches \cite{ zhang2013filtering, mongus2012parameter, mongus2013computationally, ozcan2016lidar}. Most of these methods assume that the terrain is smooth and continuous with a large height difference between neighbouring points on ground and non-ground objects. Therefore, the performance of these methods often decreases through wrongly filtering hilly regions and large buildings.

Because of the simplicity and ease of implementation, morphology-based methods \cite{kobler2007repetitive, pingel2013improved, li2014improved, mongus2014ground} are mostly used in ground filtering. However, finding the correct structuring element size is a problem in these methods. While a small structuring element is needed for filtering points on vegetation, tree, and cars, a large structuring element should be used for filtering points on buildings.

In this paper we propose an fast ground filtering approach with an efficient DTM representation capable of preserving detailed geological features and applicable to both urban and non-urban landmasses. Unlike most other methods that try to
extract ground points via many iterations for DTM generation, the proposed technique extracts all ground points via a series of atomic operations geared towards preserving geological features and eliminating non-ground points. The main hypothesis in the proposed method is that non-ground objects produce sharp variations in elevation within a spacial neighborhood. Hence, we propose using region growing for segmenting non-ground objects. The method is tested on a number of point cloud data sets obtained the United States Geological Survey. The proposed method is also compared with the existing methods. 

\section{The Proposed Approach}
\label{sec:techapp}

\begin{figure}[ht]
  \centering
  \includegraphics[width=\linewidth]{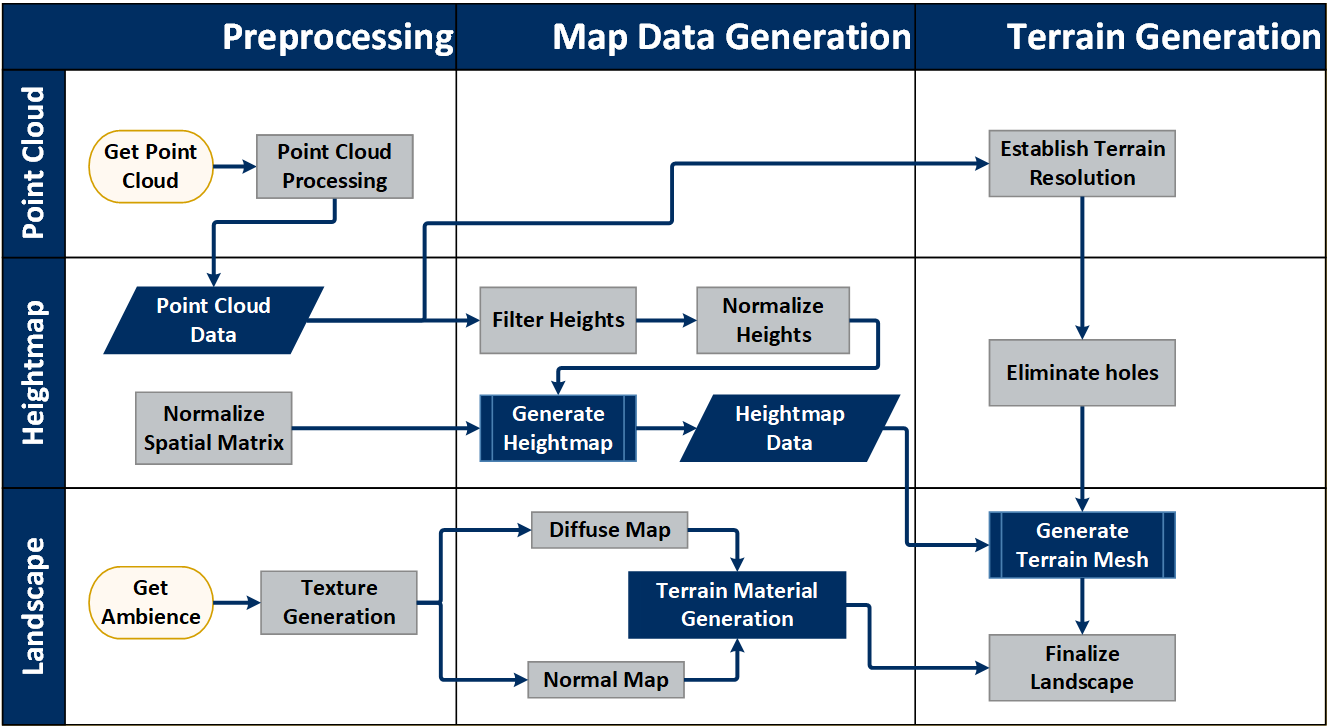}
  \caption{The proposed processing pipeline.}
  \label{fig:techApproachFlowchart}
  \vspace{-.2in}
\end{figure}

Fig.~\ref{fig:techApproachFlowchart} shows an overview of the proposed Point Cloud Filtering and DTM generation pipeline. The proposed architecture is comprised of three components, i.e., preprocessing stage, map generation module, and terrain generation module, shown as the vertical tracks. These components in turn process three different data structures in the form of Point Clouds, Heightmaps, and Landscape mesh. 

The first stage of the proposed pipeline is the preprocessing step. In this stage the Lidar point cloud data is processed to represent a gridded topological form. To accomplish this task, we perform the nearest neighbor interpolation in conjunction with a kernel-based statistical outlier removal to generate the raster grid from the Lidar point cloud. In this step, the three data structures representing the point cloud data, the spatial matrix of the map data, and the landscape texture and material data will be established and ready for processing. The texture data is extracted from the point clouds photometric information, if this information is available. The photometric information will be used to produce a diffuse map as well as a normal map for the terrain materials.

The second stage in the pipeline is responsible for generating the heightmap data representing a topological formation of the terrain as well as shader models for rendering a physically realistic view of the material applied on the surface of the terrain. In this stage, the topological spatial matrix is processed in order to accomplish two tasks. First, the overall topological and geological statistics of the terrain is learned by employing Singular Value Decomposition (SVD). Second, the learned statistics of the overall terrain topology is combined with the first and second order statistics of the point cloud data to eliminate the non-terrain objects while preserving details of the geological features of the terrain.

The last step in the pipeline is the terrain generation stage. In this stage, the resolution of the final landscape is calculated from the overall point cloud data. This information is then used to fill the holes introduced in the topological heightmap as a result of non-terrain object segmentation. Once the overall heightmap of the terrain is established, the terrain mesh generation process will generate an efficient digital mesh model for the terrain represented at various Level of Detail (LoD) information. At this stage the shader models for the terrain materials are also computed and applied to render the terrain.

\subsection{The Preprocessing Step}
The preprocessing stage of the proposed pipeline, shown in Algorithm~\ref{alg:DTMPreProcAlg}, is responsible for initializing the gridded and rasterized data structures for the Terrain heightmaps, texture maps, and shader materials. 

\vspace{-.15in}

\begin{algorithm}[ht]
    \DontPrintSemicolon
    \KwData{$P:$ Input-Point Cloud.  // point $P_i = (x_i,y_i,z_i,r_i,g_i,b_i)$}
    \KwResult{\\$L:$ Landscape Point Cloud Data File.\\$T:$ Landscape Texture.\\$M:$ Landscape Layered Material.}
    \Begin{
    $L(x,y) \leftarrow$ Stat\_Outlier($P, th$) // Eq.(\ref{eq:StatisticalOL}) \; 
    \For{all $P_i$}{
    Find $L_l(x,y,z)$ lowest and $L_h(x,y,z)$ highest Lidar Returns Eq.(\ref{eq:LhLl})\;

}
    Set texture Coordinates:
        \Begin{
        $T(u,v)=new\_Texture(u,v)$  // coordinate map from Eq.(\ref{eq:texmapcoord})\;
        }
        Set Shader Material:
        \Begin{
        $M\leftarrow new\_Material(Diff(u,v),Norm(u,v))$
        }
}
    \caption{Preprocessing Stage of the DTM Generation Pipeline.}\label{alg:DTMPreProcAlg}
\end{algorithm}

\subsubsection{Statistical Outlier Removal:}
The first step in cleaning out the input Lidar point-cloud data is to eliminate outliers. Outliers include points introduced to the point cloud due to noise or small moving objects, such as airplanes, located at drastically different heights than the terrain, need to be eliminated. In order to perform this task, we first build a non-parametric density estimation of the point cloud in a local spatial neighborhood \cite{tavakkoli2005automatic}.

Assuming an outlier threshold of $th$, we eliminate points from the point cloud data whose probability of belonging to the known distribution from which the Lidar data is generated falls below $th$. This probability is calculated using the non-parametric kernel density estimation, below:
    \begin{equation}\label{eq:StatisticalOL}
        P(z_i|inlier) = \frac{1}{|N_k (z_i)|}\sum\limits_{z_j \in N_k (z_i)} \frac{1}{\sigma\sqrt{(2\pi)}}exp\left(- \frac{\sigma(p_j ,p_i)}{2h^2}\right)  
    \end{equation}
    
where $z_i$ is the height value of the $i$th point $p_i$ in the point cloud data $p_j$ at a spatial neighborhood location of $N_k\in\mathbb R^2$.

\subsubsection{Top and Bottom Returns:}
Once the statistical outliers are eliminated, we will need to determine the most likely ground points. In order to accomplish this task, we will set two rasterized data structures for the lowest return and the highest return points at a location $(x,y)$ denoted as $L_l(x,y,z)$ and $L_h(x,y,z)$: 
\begin{equation}\label{eq:LhLl}
    \forall P_i 
    \qquad
    \begin{bmatrix} 
        L_l(x,y,z)\\
        L_h(x,y,z)
    \end{bmatrix} =
    \begin{bmatrix}
       \min\limits_{z_i}( P_i: x=x_i \& y=y_i)\\
       \max\limits_{z_i}(P_i: x=x_i \& y=y_i)
    \end{bmatrix}
\end{equation}
where $x\in\left[min(x_i),max(x_i)\right]$ and $y\in\left[min(y_i),max(y_i)\right]$.

\subsubsection{Shader and Texture Initialization:}
In order to render physically realistic materials on the surface of the final DTM, we will establish the data structures $T(u,v)$ and $M$ as the texture map and the landscape material, respectively.
First, a mapping between the spatial domain of the point cloud $(x,y)\in\mathbb R^2$ and the texture-coordinates $(u,v)$ is determined:
\begin{equation}\label{eq:texmapcoord}
    (u,v)^T =  \begin{bmatrix}
    \Phi_u:(x_{min},x_{max})\rightarrow (0,1)\\
    \Phi_v:(y_{min},y_{max})\rightarrow (0,1)
    \end{bmatrix}
\end{equation}

Next, the shader material for the landscape is initialized based on the photogrametric information, if this information is included in the Lidar point cloud data. Suppose for each point $P_i$ in the point cloud data, the photogrametric information is given in the form of $C_i= (r_i,g_i,b_i)$ color components. The details about the computation of the diffuse and normal channels of the landscape material are discussed later in the paper in section~\ref{sec:TDMmodling}.

\subsection{The Heightmap Generation Step}
Once the point cloud data is refined during the pre-processing step, it is passed through the heightmap generation stage of the algorithm to produce a two-dimensional structure maintaining the overall height associated with the terrain surface. This heightmap object is then utilized to generate a three-dimensional model of the terrain surface as a 3D mesh object. This section discusses the process of generating the terrain heightmap by removing non-terrain objects while preserving significant geological features.

\begin{algorithm}[ht]
    \DontPrintSemicolon
    \KwData{\\$L_l,L_h:$ Landscape Top and Bottom Point Cloud Data.\\$P:$ Point Cloud Data.}
    \KwResult{\\$H:$ Landscape Heightmap Data File.}
    \Begin{
   
    \For{all $P_i$}{
    $\hat L = L_l \cap L_h$ // Non-ground overhangs \;
    $ L = (L_l\cup L_h) - \hat L $ // Potential ground points \;
    $\hat H \leftarrow L.Heights$ // Eq.(\ref{eq:H1})\;
    Find $\hat g \leftarrow$ S.V.D. $(\hat H)$ // Eq.(\ref{eq:SVDEQ1})\;
    $g\leftarrow \textbf{inPaint}(\hat g)$  // Fill holes \;
}
}
    \caption{Heightmap Generation Stage of the DTM Pipeline.}\label{alg:DTMHeightmapAlg}
\end{algorithm}

Algorithm~\ref{alg:DTMHeightmapAlg} shows the overall pipeline of generating the terrain surface heightmap. The process starts by taking the top and bottom point cloud data structures generated from the pre-processing phase to determine the potential ground points and eliminate the over-hangs. Then a polynomial function with sufficient local variance and smooth global consistency is fit onto the data to estimate the overall structure of the terrain ground. This is utilized to computer the ground heightmap values for each point in the landmass.

\begin{figure}[ht]
  \centering
    \subfigure[]{\includegraphics[height=1.5in]{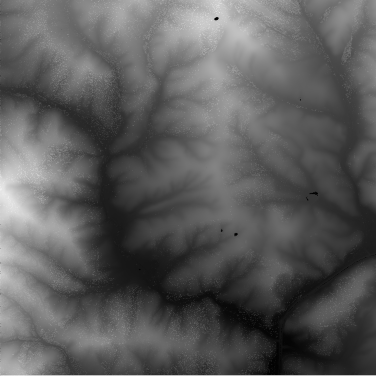}\label{fig:Original}}
  \subfigure[]{\includegraphics[height=1.5in]{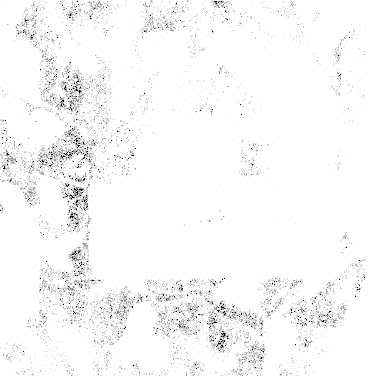}\label{fig:Non-GroundPoints}}
    \subfigure[]{\includegraphics[height=1.5in]{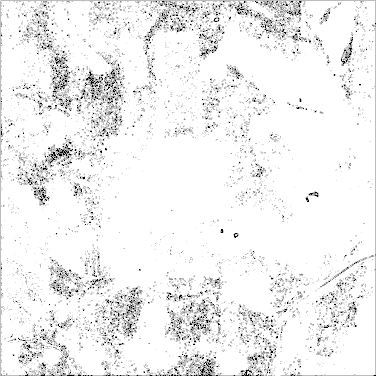}\label{fig:ElevationDiscrepancy}}
    \subfigure[]{\includegraphics[height=1.5in]{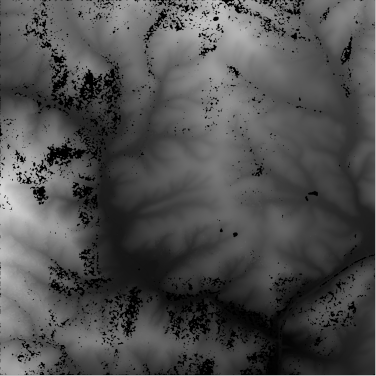}\label{fig:ElevationResolved}}
    \subfigure[]{\includegraphics[height=1.5in]{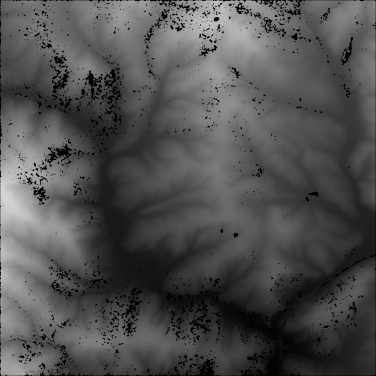}\label{fig:ElevationResolvedSteepRegionGrown}}    \subfigure[]{\includegraphics[height=1.5in]{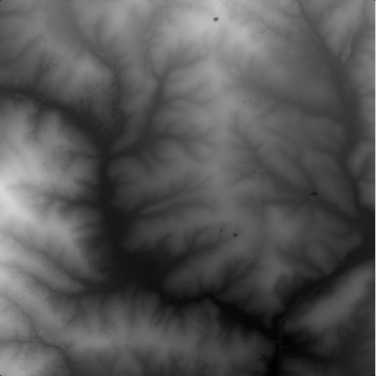}\label{fig:FinalHeightmap}}
  \caption{Terrain Modeling Heightmap Generation Step Performed on Idaho Dataset. (a) Original Data with No-terrain Elements. (b) Dark Areas are Non-Terrain LiDar Returns. (c) Dark Areas are Non-Terrain Geological Features. (d) Non-Terrain Geological Features are Removed. (e) Non-Terrain Areas are Removed. (f) Final Terrain Heightmap.}
  \label{fig:HeightmapSteps}
  \vspace{-.25in}
\end{figure}

\subsubsection{Terrain Height Estimation}
With the Lowest $L_l$ and the highest $L_h$ LiDar returns from the point cloud data, we start modeling the heightmap of the terrain. Each point $(x,y,z)$ in a point cloud belongs to one of two classes, i.e., the ground or the non-ground objects. Both $L_l$ and $L_h$ are quantized in such a way as to represent 2D grids ranging from $(x_{min}, y_{min})$ to $(x_{max}, y_{max})$.

It is trivial to eliminate overhangs (or points covering the ground area) if both the ground position and the overhang points are visible within the point cloud data. Points within a spatial location $\mathcal{R}(x,y)$ are considered to belong to the non-ground object covering the surface of the ground if they exist in both $L_l$ and $L_h$ structures. Therefore, the first iteration of the heightmap is generate by interpolating the height values of all points in $L_l$ that do not belong to $L_h$ as:

\begin{equation}\label{eq:H1}
\hat H = h(x,y)=\frac{1}{Size(\mathcal R)}\sum_{(x,y,z)\in\mathcal R} \{z | (x,y,z)\in L\}
\end{equation}

The height of the ground in a landmass may be considered as a low-degree polynomial with the non-ground objects, e.g. shrubbery, biomass, and man-made structures, disrupting the natural curvature and geological features of the terrain. Therefore, we postulate that the heightmap of the terrain is a combination of a ground function $g(x,y)$ and an anomalous function $\mathcal N(x,y)$:
\begin{equation}\label{eq:groundFunc}
    h(x,y) = g(x,y)+\mathcal N(x,y)
\end{equation}

where $h$ is a heightmap calculated from raw point cloud data (Fig.~\ref{fig:Original}), $g$ is a low-order polynomial function with high degrees of smoothness over a large spatial area representing the ground heightmap (Fig.~\ref{fig:FinalHeightmap}), and $\mathcal N$ represents the non-ground geological and man-made features shown in Fig.~\ref{fig:Non-GroundPoints} and Fig.~\ref{fig:ElevationDiscrepancy}.

This formulation represents the terrain heightmap modeling as a novelty detection question \cite{tavakkoli2009novelty}. Therefore, we represent the ground region of the terrain heightmap data $g$ as a polynomial with degree $N$ of the following form:
\begin{equation}\label{eq:GroundEQ}
    g(x,y) = \sum_{i,j = 0}^{N}a_i b_j (x^i \cdot y^j)
\end{equation}
Using the above formulation, and given the heightmap from Eq.(\ref{eq:H1}), we need to solve the linear system of equations resulting from all $(x,y)$ values of $\hat H$ as follows:
\begin{equation}\label{eq:SVDEQ1}
    \left[h(x_1,y_1) \cdots h(x_N,y_N)\right]^T = 
    \begin{bmatrix}
    a_0b_0&a_1b_0&\cdots &a_Nb_N \\ \vdots &\vdots&\ddots &\vdots\\a_0b_0&a_1b_0&\cdots &a_Nb_N
    \end{bmatrix} 
    \begin{bmatrix}1 & x & y &  \cdots & x^Ny^N \end{bmatrix}
\end{equation}
% The above system of equations is solved using Singular Value Decomposition (SVD) \cite{golub1971singular}. The solution to the above equation lays in the null-space of the transformation formed by SVD. Once the coefficients in Eq.(\ref{eq:SVDEQ1}) are computed, we generate the terrain function using these coefficient. 

This terrain function may be visualized as the combination of Fig.~\ref{fig:ElevationResolved} and Fig.~\ref{fig:ElevationResolvedSteepRegionGrown}, in which the darker areas represent non-ground objects encoded as $\mathcal N$. These dark areas produce holes in the ground heightmap and are filled using an automatic inpainting algorithm similar to \cite{van2016filling}. The final terrain heightmap is shown in Fig.~\ref{fig:FinalHeightmap}.

\begin{figure}[ht]
  \centering
    % \subfigure[]{\includegraphics[width=2.5in]{images/ID-PointCloudData}\label{fig:OriginalPointCloud}}\\
  \subfigure[]{\includegraphics[width=3in]{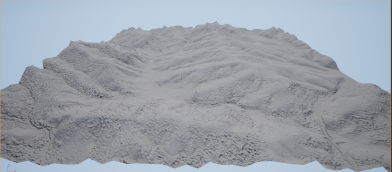}\label{fig:OriginalTerrain}}
  \subfigure[]{\includegraphics[width=3in]{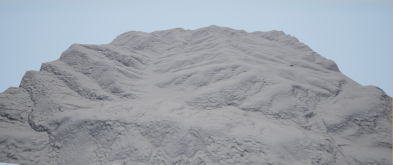}\label{fig:DetectedTerrain}}
  \caption{Terrain meshes: (a) Terrain mesh from the original point cloud. (b) Terrain mesh with the proposed heightmap generation technique.}
  \label{fig:ResultsHeightmap}
  \vspace{-.1in}
\end{figure}

\subsection{The Terrain Modeling Step}\label{sec:TDMmodling}
Digital Terrain Models are employed in a number of applications ranging from geographical analysis, biomass and environmental studies, etc. In order for a DTM to be useful for its intended application, it must be generated in such an efficient manner as to allow for realistic rendering, interactivity, and efficient manipulation. To this end, we propose the use of the Unreal Engine 4's Landscapes~\cite{tavakkoli2018game}. Algorithm~\ref{alg:DTMShaderAlg} provides an overview of this stage of the pipeline responsible for generating the 3D mesh of the terrains as well as shader materials employed for physically realistically rendering of the terrain.

\begin{algorithm}[t]
    \DontPrintSemicolon
    \KwData{\\$P:$ Point Cloud Data \\$H:$ Heightmap}
    \KwResult{\\$M:$ Terrain Mesh. $H:$ Terrain Heightmap. $T:$ Terrain Texture. $Mat:$ Shader Material.}
    \Begin{
    Mesh.Vert: M.(vertex.x,vertex.y)$\xleftarrow{\Phi}(H.x,H.y)$
    $M\leftarrow \textrm{Interpolate}[h(\textrm{Grid}(x,y))]$\;
    \For{all $P_i$}{
            $T\leftarrow$PC2Texture$(P,H)$  // Generate Texture from Point Cloud\;

}
    Calculate Terrain Extent\;
    $(U,V)\leftarrow$ Texture Coordinate Mapping\;
    $Mat\leftarrow (Diffuse_{uv},Normal_{uv})$ // Shader Program\;
    
}
    \caption{Mesh Generation and Shader Programming of the DTM Pipeline.}\label{alg:DTMShaderAlg}
\end{algorithm}

\vspace{-.15in}
\subsubsection{Terrain Mesh Modeling:}
\label{ssec:Mesh}
The 3D mesh representing the ground surface is generated by applying a polygonal mesh based on the heightmap generated from the previous step. In this stage of the algorithm, a 2D grid is generated for each pair of $(x,y)$ coordinates associated with pixels in the heightmap.

\begin{figure}[ht]
  \centering
    \subfigure[]{\includegraphics[height=1.5in]{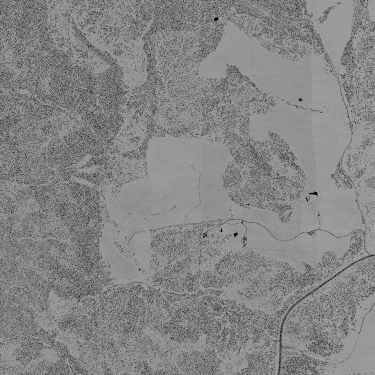}\label{fig:OriginalTex}}
  \subfigure[]{\includegraphics[height=1.5in]{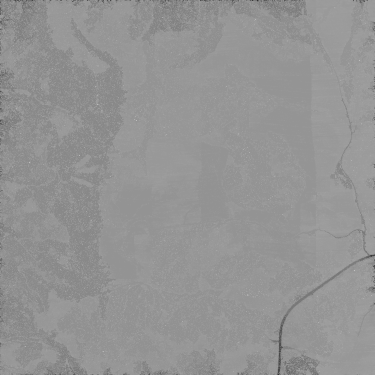}\label{fig:SparseTex}}
    \subfigure[]{\includegraphics[height=1.5in]{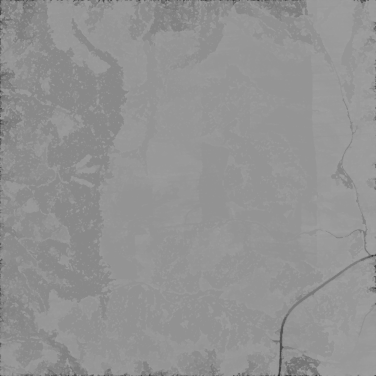}\label{fig:FinalTex}}
\caption{Terrain Texture: (a) Original Sparse Texture. (b) Dense Texture. (c) Modified Dense Texture.}
  \label{fig:Textures}
\end{figure}

\subsubsection{Terrain Texture Modeling:}\label{ssec:Texture} With the mapping between the heightmap data and the point cloud data established, an interpolation technique is used to sample the color (or intensity) values from point clouds in a neighborhood that map onto the terrain mesh object. Fig.~\ref{fig:Textures} shows the resutls of the interpolation steps taken to generate a photorealistic texture for the terrain from the Point Cloud data.

\begin{figure}[ht]
  \centering
    \includegraphics[width=4.2in]{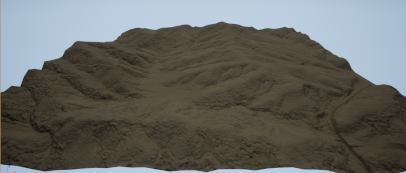}  
  \caption{Final Terrain Model with Shader Parameters Applied.}
  \label{fig:FinalMaterialResults}
\end{figure}

\subsubsection{Terrain Material Modeling:}\label{ssec:Material}
The material applied to the surface of the terrain mesh is comprised of two main channels, a diffuse channel and a normal channel. The diffuse channel of the material utilizes the texture coordinates to map the color (or intensity) values of the terrain texture on the surface of the 3D terrain mesh. The normal channel is computing using a normal map generation technique \cite{gimpnormalmap}.

Fig.~\ref{fig:FinalMaterialResults} shows the final rendering of the Digital Terrain Model with the material applied. As it can be seen, the quality of the rendering is quite realistic. Note the various geological features preserved, while the man-made structures or non-ground objects are effectively removed from the terrain model. The texture applied on the surface of the terrain in the form of a physically-based material drastically enhances the visualization of the DTM.

\section{Experimental Results}\label{sec:results}
\begin{figure}[ht]
  \centering
    \subfigure[]{\includegraphics[height=1.8in]{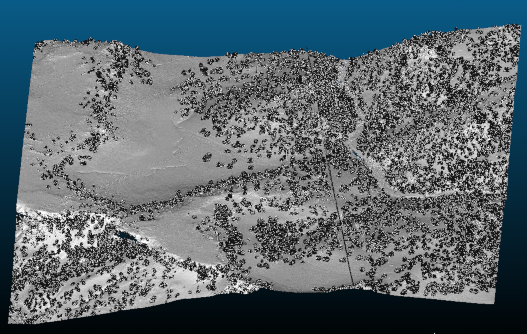}\label{fig:pointcloud}}
  \subfigure[]{\includegraphics[height=1.8in]{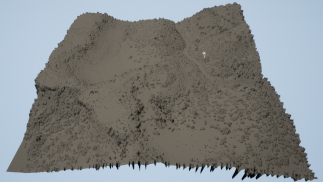}\label{fig:Reslandscape}}
  \caption{The results of the proposed framework. (a) The original point cloud data rendered in Cloud Compare software. (b) The 3D landscape DTM generated by the proposed framework and rendered in Unreal Engine 4.}
  \label{fig:results}
\end{figure}

This section presents the results of the proposed DTM technique performed on a variety of point cloud data from the USGS datasets. The first set of results (Fig.~\ref{fig:results}) demonstrates that quality of the modeled DTM compared to the rendering of the point cloud data. The point cloud data rendered in the Cloud Compare software is shown in Fig.~\ref{fig:pointcloud}. The main issue is the lack of discrimination between points belonging to the ground surface and other structural elements. The proposed technique has the ability to eliminate the non-terrain elements while preserving significant geological features as evident from Fig.~\ref{fig:Reslandscape}.

\begin{figure}[ht]
  \centering
  \subfigure{\includegraphics[height=1.85in]{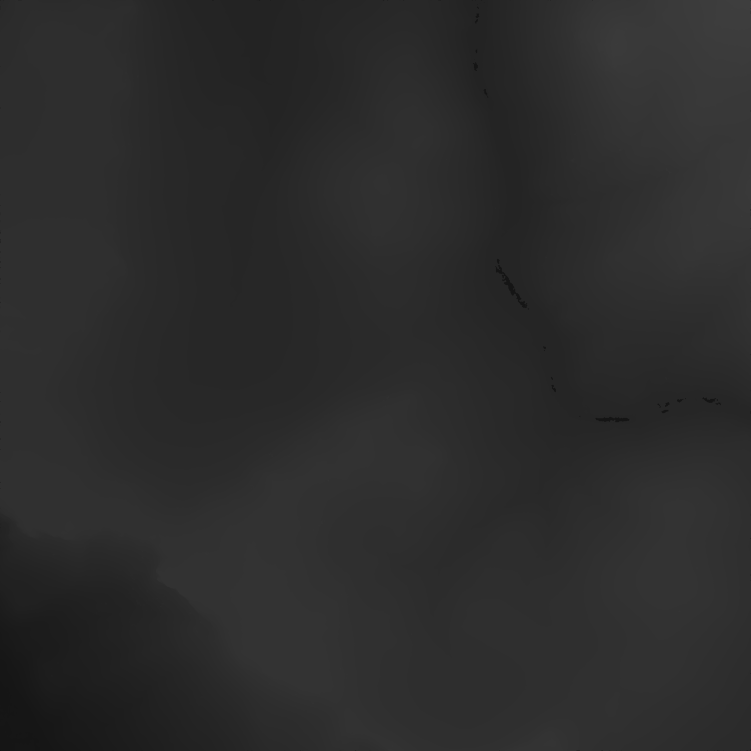}}
  \subfigure{\includegraphics[height=1.85in]{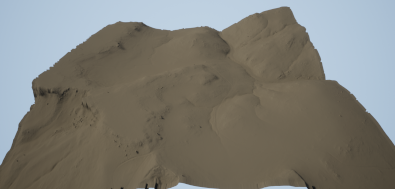}}\\
  \subfigure{\includegraphics[height=1.85in]{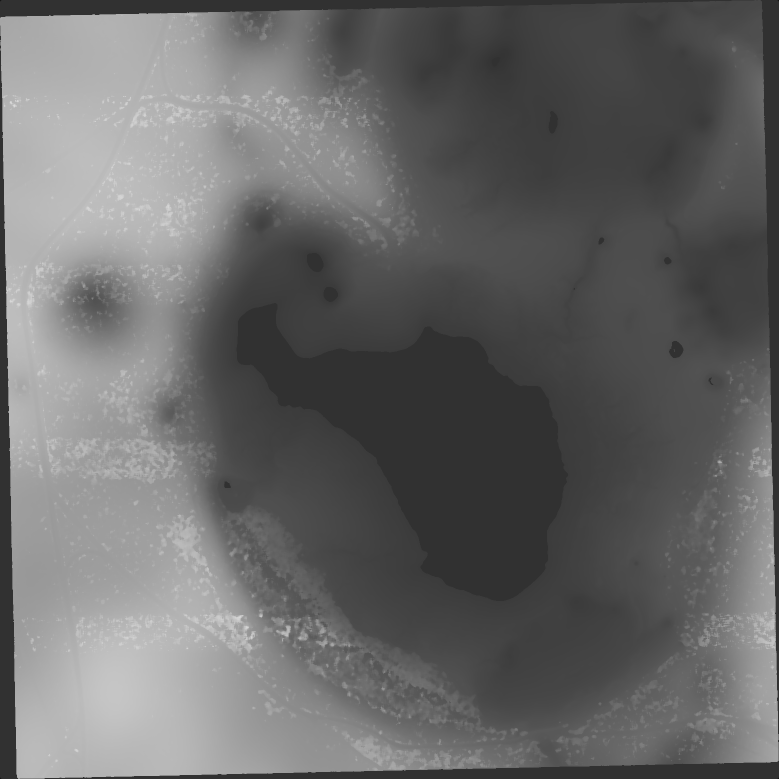}}
  \subfigure{\includegraphics[height=1.85in]{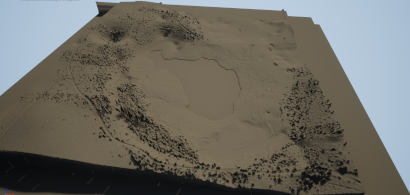}}\setcounter{subfigure}{0}\\
  \subfigure[]{\includegraphics[height=1.85in]{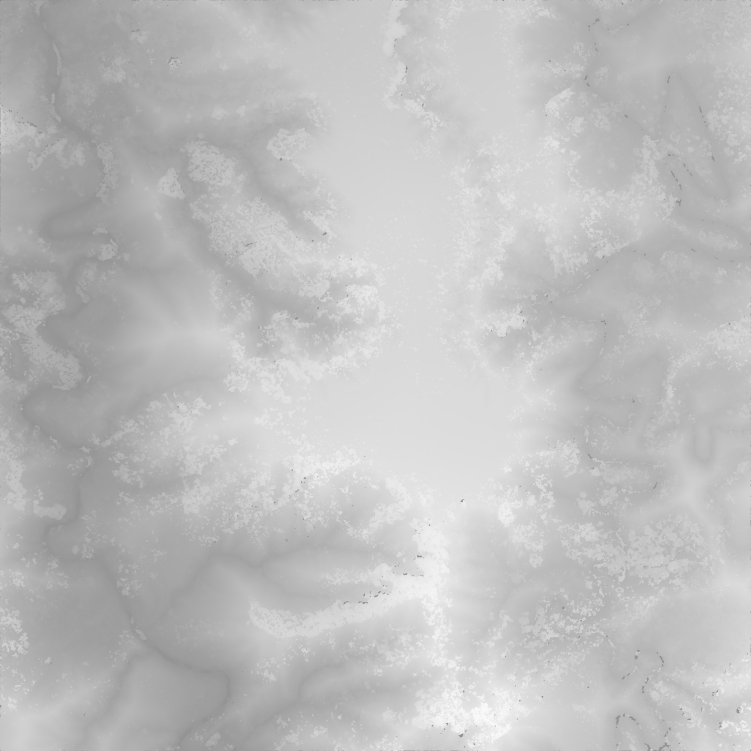}\label{fig:HMap}}
  \subfigure[]{\includegraphics[height=1.85in]{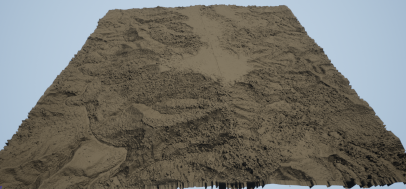}\label{fig:Landscape}}
  \caption{Heightmaps (a) and their associated terrains (b) generated by the proposed framework. From top: California Calaveras-Tuolumne (CA), Washington County (FL), and Oahu (HI), respectively.}
  \label{fig:multipleResults1}
\end{figure}

Fig.~\ref{fig:multipleResults1} shows the generated heightmaps (Fig.~\ref{fig:HMap}) and the 3D landscape mesh associated with each heightmap (Fig.~\ref{fig:Landscape}. As seen from the figures, the proposed DTM mesh objects represent the geological features quite accurately.

% \begin{figure}[ht]
%   \centering
%     \subfigure[]{\includegraphics[height=1.3in]{CA_Tuolumne_Terrain_LandscapeCloseup.PNG}\label{fig:VRcloseup}}\hfill
%   \subfigure[]{\includegraphics[height=1.3in]{CA_Tuolumne_Terrain_Closeup.PNG}\label{fig:Pointcloudcloseup}}
%   \caption{The closeup comparisons of the results of the proposed framework and the current point cloud viewing systems. As it can be seen from the generated VR landscape (a), the fidelity of visual representation of terrain features are much higher compared to original point cloud representation of the same area (b).}
%   \label{fig:CloseupResults}
% \end{figure}

% \begin{figure}[t]
%   \centering
%   \subfigure{\includegraphics[height=1.32in]{GA_OkefenokeeHMap16_1k_Floor_Med3x3_G1_5}}\hfill
%   \subfigure{\includegraphics[height=1.32in]{VR_GA_OkefenokeeHMap16_1k_Floor_Med3x3_G1_5}}
%   \subfigure{\includegraphics[height=1.32in]{VA_ChesapeakeBaySouthHmap16_1k_Floor_Med3x3_HP_G1_5}}\hfill
%   \subfigure{\includegraphics[height=1.32in]{VR_VA_ChesapeakeBaySouthHmap16_1k_Floor_Med3x3_HP_G1_5}}\setcounter{subfigure}{0}
%   \subfigure[]{\includegraphics[height=1.35in]{WA_mtBakerHmap16_1k_Floor_Med3x3_HP_G1_5}\label{fig:HMap2}}\hfill
%   \subfigure[]{\includegraphics[height=1.35in]{VR_WA_mtBakerHmap16_1k_Floor_Med3x3_HP_G1_5}\label{fig:Landscape2}}
%   \caption{Heightmaps (a) and their associated terrains (b) generated by the proposed framework. From top: Okefenokee (GA), Chesapeake Bay (VA), and Mount Baker (WA), respectively.}
%   \label{fig:multipleResults2}
% \end{figure}

\section{Conclusions and Future Work}
In this paper we proposed a pipeline for generating Digital Terrain Models (DTM) from a variety of Lidar-based point cloud datasets. The proposed pipeline automatically generates heightmaps by eliminating non-ground points from the point cloud and interpolating the surface height values from the remaining points. The texture and materials are also created to provide photorealistic rendering of the terrain 3D mesh. There are a number of future directions to this work. Performing semantic segmentation on the 3D point cloud data may add higher level information to the data useful for effective generation of heightmaps.
\bibliographystyle{unsrt} 
\bibliography{ISVC19-RemoteSensing/references}

\end{document}